\def\year{2019}\relax
\documentclass[letterpaper]{article} 
\usepackage{aaai19}  
\usepackage{times}  
\usepackage{helvet} 
\usepackage{courier}  
\usepackage[hyphens]{url}  
\usepackage{graphicx} 
\usepackage{graphicx}  
\usepackage{booktabs}
\usepackage{amsmath}
\usepackage{amssymb}
\usepackage{dsfont}
\usepackage{mathtools}
\usepackage{booktabs}
\usepackage{siunitx}
\usepackage{xcolor}
\newcommand{\citet}[1]{\citeauthor{#1} \shortcite{#1}}
\title{Interpretable Image Recognition with Hierarchical Prototypes}

\author{\textbf{Peter Hase,$^{1, \hspace{.5mm} 2}$ Chaofan Chen,$^2$ Oscar Li,$^2$ Cynthia Rudin$^2$}\\
\textsuperscript{1}UNC Chapel Hill\\
\textsuperscript{2}Duke University\\
peter@cs.unc.edu, \{chaofan.chen, rl144, cynthia.rudin\}@duke.edu
}

\begin{document}

\maketitle

\begin{abstract}
Vision models are interpretable when they classify objects on the basis of features that a person can directly understand. Recently, methods relying on visual feature prototypes have been developed for this purpose. However, in contrast to how humans categorize objects, these approaches have not yet made use of any taxonomical organization of class labels. With such an approach, for instance, we may see why a chimpanzee is classified as a chimpanzee, but not why it was considered to be a primate or even an animal. In this work we introduce a model that uses hierarchically organized prototypes to classify objects at every level in a predefined taxonomy. Hence, we may find distinct explanations for the prediction an image receives at each level of the taxonomy. The hierarchical prototypes enable the model to perform another important task: interpretably classifying images from previously unseen classes at the level of the taxonomy to which they correctly relate, e.g. classifying a hand gun as a weapon, when the only weapons in the training data are rifles. With a subset of ImageNet, we test our model against its counterpart black-box model on two tasks: 1) classification of data from familiar classes, and 2) classification of data from previously unseen classes at the appropriate level in the taxonomy. We find that our model performs approximately as well as its counterpart black-box model while allowing for each classification to be interpreted.

\end{abstract}


\section{Introduction}

\begin{figure}[h!]
  \centering
  \includegraphics[width=0.46\textwidth]{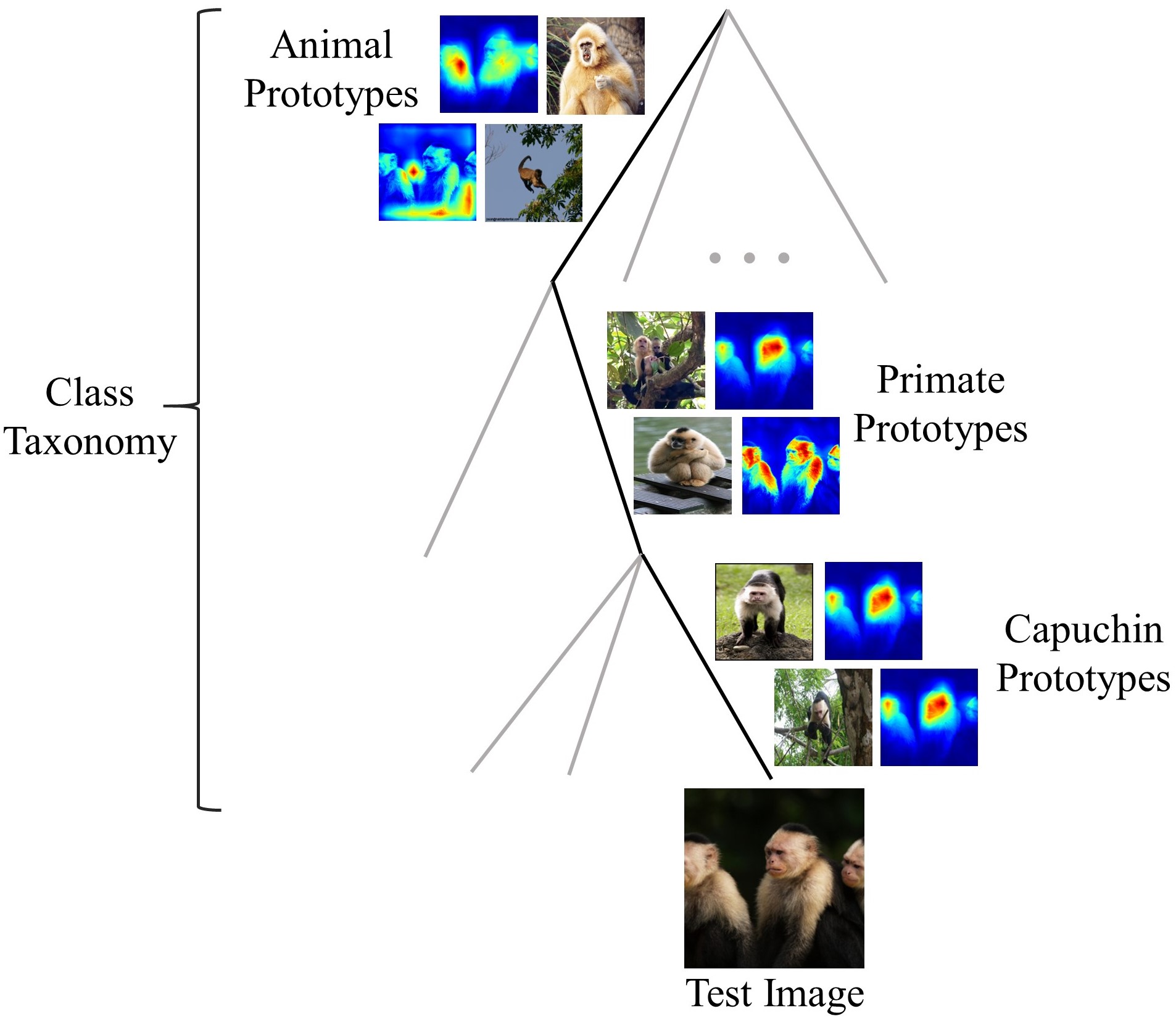} 
  \caption{The capuchin shown here is classified at three levels in a pre-defined taxonomy: animal, primate, and capuchin. The classification is made based on similarities between the latent representation of the capuchin and learned prototypes corresponding to each of these three hierarchically organized classes. Beside each prototype, there is a heat map showing the localized areas in the test image that highly activated the prototype. Hence, we see the parts of the image that led the model to classify it as an animal (rather than, e.g., a vehicle), a primate (rather than a non-primate), and finally as a capuchin (rather than an orangutan or gibbon). For the full class taxonomy, see Figure \ref{fig:taxonomy}.}
  \label{fig:lead}
\end{figure}

What is clear from the study of human vision is that we organize the world into ``inductively rich" categories that relate to each other taxonomically: we glean useful information from recognizing that something is an animal, and we draw even more useful information from seeing that it is a tiger rather than a cat \cite{Bloom}. Further, we can explain our visual judgments by pointing to prototypical features that an object possesses as evidence for its membership in the class to which those features correspond \cite{oneshothierarchy}; a certain animal is a tiger because it is a large cat with black stripes, approximately orange fur, menacing teeth, etc. Using some internal taxonomy, people can explain their visual judgments at multiple levels of abstraction, and these explanations may differ across each level. What distinguishes animals from vehicles is different from what distinguishes lions and tigers, for instance. People can also tell when an object resides within a coarse class (like animals) but does not belong to any familiar fine class (like deer). That is, we can tell when something is a kind of animal we have never seen before. 

For computer vision models to replace human judgment on important image recognition tasks, these models should fulfill the same functions and exhibit the same transparency as we do, which is to suggest that vision models should 1) classify images on the basis of human-interpretable features such as visual feature prototypes, 2) predict image classes not just at the level of the dataset labels, but also at each level of a taxonomy that is known to organize the classes, and 3) detect when images belong to never-before-seen subclasses within some coarse class in the known taxonomy. 

Why add a hierarchical class structure and novel class detection to an interpretable model? First, such an approach makes explicit the trade-off between information gain and accuracy, since it is easier to distinguish among objects at less informative levels of a taxonomy \cite{hedgingyourbets}. Users can elect to make their decision using only the more reliable but more coarse-grained classification. This option is useful when policy responses to a prediction do not change after a certain level of specificity, or when they differ between two ``sibling" classes. For example, a decision-maker who is unsure whether an object is a pistol or an assault rifle might nonetheless produce an appropriate policy response simply by virtue of knowing the object is a gun.

Second, explanations for predictions can be tailored to their corresponding taxonomical level. We may identify the reasons for an ambulance being an automobile (which could include the presence of wheels), and then for it being an ambulance (which could include the presence of sirens, not simply the presence of wheels). This helps us focus our understanding of each image's prediction to the most specific level at which we wish to distinguish its class from other classes. Further, when the predictions are wrong, we can see at what level of specificity they went wrong.

Lastly, this approach enables the interpretation of predictions for images from novel (never-before-seen) classes, as long as these novel classes fall under other broader classes in the model's known taxonomy. Vision models deployed in live environments will inevitably encounter such images, and it will be useful for them to recognize both that these images belong to novel classes and that they are instances of some familiar but more coarse-grained class. 

In this paper, we present an algorithm that performs the three functions described above. In doing so, we draw upon work from three frameworks in computer vision: 1) interpretability through feature prototypes, 2) hierarchical class organization, and 3) novel class detection. Our algorithm combines and builds upon insights from each framework, allowing for the interpretable classification of images at multiple levels of taxonomical specificity, even when these images come from novel classes (that fall under a broader class in the known taxonomy). With a subset of ImageNet, we test our model against its counterpart black-box vision model on two tasks: classification of data from familiar classes and classification of data from previously unseen classes at the appropriate level in the model taxonomy. We find that our model performs approximately as well as its counterpart black-box model while producing interpretable predictions. 

We report several accuracy metrics here, including 1) fine-grained accuracy on in-distribution data, 2) coarse-grained accuracy on novel data, and 3) novel class detection accuracy. We also give a quantitative evaluation of the quality of our interpretable model's learned latent space.\footnote{We are making our code publicly available at: \\ \url{https://github.com/peterbhase/interpretable-image}.}

\section{Related Work}

\begin{figure*}[ht]
  \centering
    \includegraphics[width=0.7\textwidth]{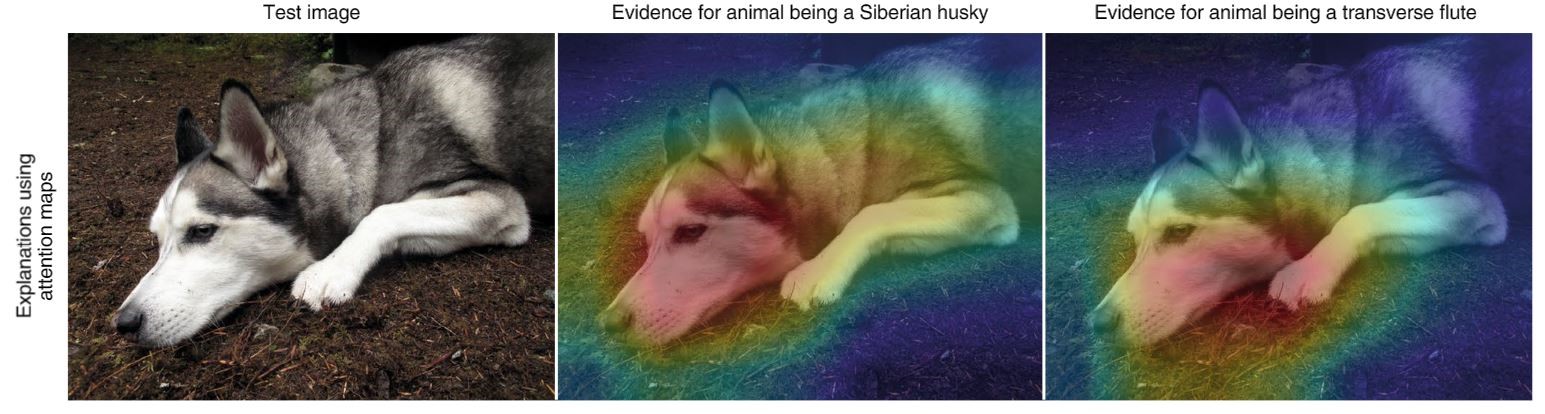}
  \caption{Saliency maps show where the model is looking, but they do not tell \textit{why} the model classifies an image as it does. Our prototype-based model provides more localized features, leading to better explanations of model classifications (see Figure \ref{fig:case_study1}).}
  \label{fig:please_stop}
\end{figure*}

Within computer vision, there are long lines of prior research in each of hierarchical classification, interpretable modeling, and novel class detection. To date, however, the approaches have not been successfully combined. 

\subsubsection{Interpretable Models.} There is no shortage of \textit{post-hoc} interpretations of CNN-based vision models \cite{Erhan,lee2009convolutional,simonyan2013deep,pmlr-v70-sundararajan17a}, but there are fewer methods where an attempt is made to learn explicitly interpretable features. A few identify subsections of an image that were important to a classification, e.g., the class-attention maps of \citet{pinheiro2015image} and \citet{zhou2016learning}. Others feed only a localized portion of the image to the model that is selected in a supervised manner, with densely labeled data \cite{huang2016part,zhou2018interpretable}, or in an unsupervised manner with auxiliary networks pre-trained for this purpose \cite{simon2015neural}. The shortcomings of class attention and saliency-based approaches in particular are exemplified in Figure \ref{fig:please_stop}. 

The most interpretable methods include the prototype-based approaches of \citet{branson2014bird}, \citet{LiLiuChenRudin}, and \citet{thislookslikethat}. Our approach differs from each along several important dimensions. Using the prototype-specific heat map method from \citet{thislookslikethat}, we show that our prototypes encode for local information (i.e. \textit{image parts}), a quality of prototypes that \citet{branson2014bird} do not demonstrate for their model. \citet{LiLiuChenRudin} use a decoder to visualize prototypes of MNIST classes, which does not work for complex naturalistic images. 

Our approach differs from \citet{thislookslikethat} by organizing the prototypes hierarchically rather than in a ``flat" manner. Whereas past work involves learning prototypes particular to each class in the dataset, our method learns both an analogous set of prototypes, which correspond to the leaf nodes in the class taxonomy, and additional sets of prototypes for each related group of classes, which correspond to parent nodes in the class taxonomy. This enables us to interpret image classifications at each parent node in taxonomy, e.g., what makes a panda an animal rather than a vehicle as well as what distinguishes the panda from a lion. 

\subsubsection{Hierarchical Classification.}
Hierarchical classification has been performed with SVMs \cite{hedgingyourbets}, Bayesian graphical models \cite{oneshothierarchy}, CNNs, \cite{branchconnect,yolo9000,bcnn,integratingmultilevel,hdcnn}, and the use of a CNN and RNN together \cite{CNNRNN}. Typically the problem is entirely supervised, but inferring the tree structure of classes has been done in an unsupervised fashion as well \cite{learningconcepttaxonomies}. Our work falls into the supervised CNN category. Within this category, some approaches use a single CNN and construct predictions over the class taxonomy from a single network output \cite{yolo9000}, while others branch their networks to produce representations unique to each sub-classification task \cite{bcnn,hdcnn}. Our work is an instance of the latter, as our network branches at a particular point before any classifications are made.

Our approach departs from previous work in hierarchical classification through our introduction of prototypes that encode for image parts, allowing for model classifications to be directly interpreted. None of the prior CNN-based approaches make use of prototypes in the latent space; they introduce hierarchical class labeling chiefly for purposes of increasing accuracy or dealing with labels of varying specificity. There is a Bayesian graphical model that utilizes prototypes \cite{oneshothierarchy}, but the prototypes in this model are in pixel space and each represents an entire class, while our prototypes reside in a latent space and represent parts of a class. 

\subsubsection{Novel Class Detection.}
Lastly, \citet{digit6} review a variety of novel class detection methods for vision models, which are known variously as out-of-distribution detection, outlier detection, or novelty detection methods. These methods include both unsupervised and supervised approaches and predominately operate by using information from the logits that a model outputs. Unsupervised methods here rely on standard statistical outlier detection techniques \cite{opensetdeepn}, and they perform consistently worse than supervised techniques. The supervised approaches make use of classifiers on a model's logits, while of course requiring that some data are withheld from the modeling process to serve as out-of-distribution data.

We extend this body of work by adapting past methods to the context of hierarchical class organization. We introduce a method for solving the problem of detecting when an instance resides within a known coarse class (like animals) but not within any of the known sub-classes (like panda or lion).

\section{Problem Descriptions}

We describe the three problems treated by our approach.

\subsubsection{Ensuring Interpretability.} Interpretable vision models classify images on the basis of directly interpretable features. From the work of \citet{Bloom}, we identify two paths toward ensuring that features are interpretable. First, one could produce features corresponding to object properties like redness or having legs. Second, one could produce features from measures of similarity between new instances and representative instances of each class. Note that as a point of terminology, Bloom identifies the first of these approaches as the \textit{prototype} approach, while he identifies the second as the \textit{exemplar} approach. The model we introduce here is best considered to follow the prototype approach, since the model learns features that are image \textit{parts} rather than entire instances of classes. Our model could also be described as following case-based reasoning, since even as the model learns feature prototypes that represent abstract object properties, the feature prototypes are always drawn from concrete instances in the training data. 

Can we ensure the model learns \textit{meaningful} features? Quantitative metrics for interpretability have been developed \cite{bau2017network}, but they rely on particular densely labeled datasets that still may not capture all of the relevant, meaningful features for a distinct setting of deployment. Consequently, domain experts must check that features are meaningful for applications in their domain. For the domains like those captured in the ImageNet data used in this paper, a layperson can check if features are meaningful; they are not forced to trust a black box.

\begin{figure}[t]
  \centering
  \includegraphics[width=0.44\textwidth]{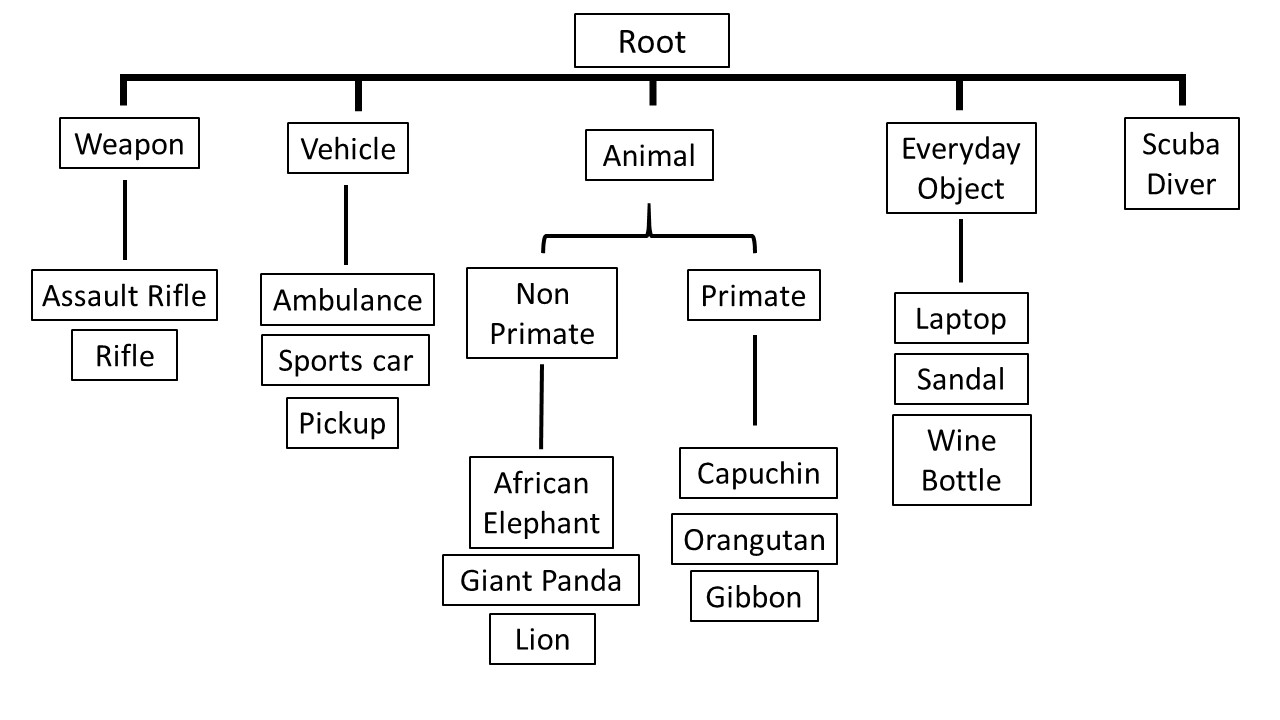}
  \caption{Our class taxonomy defined over a subset of 15 ImageNet classes, where each of the fifteen classes is represented as a leaf node.}
  \label{fig:taxonomy}
\end{figure}

\subsubsection{Hierarchical Classification.}
The task of hierarchical classification is to predict an image's class at each level of a taxonomy (i.e. tree). Suppose we have sample images from the data space $\mathcal{X}$ and labels from the space $\mathcal{Y}$. The key difference with a standard classification framework is that each label $\mathbf{y}_i$ has $k$ elements, with $\mathbf{y}_i^{(k)}$ representing an image's label at the $k^\text{th}$ level in the tree. 

Here, $\mathcal{Y}^{(0)}$ denotes the root, $\mathcal{Y}^{(1)}$ will represent the first set of children below the root, which correspond to the coarsest classes, and $\mathcal{Y}^{(K)}$ will represent the finest level. We seek to learn a function $f: \mathcal{X} \rightarrow \mathcal{Y}$ that approximates the true distribution $P(\mathcal{Y}|\mathcal{X})$ over paths in the tree; each path corresponds to an image's full label, e.g. \{animal, cat\}. Physically impossible paths, like \{animal, truck\} are known a priori to have $0$ probability. Not all branches need to be the same depth, though for notational convenience we will always write label sequences through $\mathcal{Y}^{(K)}$.

This task is accomplished by learning each of the conditional distributions within a factorization of the full joint distribution $P(\mathcal{Y} |\mathcal{X}) = P(\mathcal{Y}^{(1)},...,\mathcal{Y}^{(K)}|\mathcal{X}) $.  

Then there are as many distributions to learn as there are parent nodes, counting one root node corresponding to the distribution over $\mathcal{Y}^{(1)}$. Each distribution $P(\mathcal{Y}^{(k+1)}|\mathcal{Y}^{(k)} = c^{(k)}, \mathcal{X})$ represents the multinomial distribution over the children classes of the parent node $c^{(k)}$ on the $k^{\text{th}}$ tree level, where $c^{(0)}$ is the class of all known entities: 
\begin{align*}
P(\mathcal{Y}|\mathcal{X}) &= P(\mathcal{Y}^{(1)} | \mathcal{X}) \times  ... \times P(\mathcal{Y}^{(K)}| \mathcal{Y}^{(K-1)}, \mathcal{X}),
\end{align*}
which is a typical objective for multinomial classification. 

\subsubsection{Novel Class Detection.}
Novel class and out-of-distribution detection methods provide a mechanism for identifying data not from a known distribution. This can either be considered an unsupervised problem, or a binary classification problem, if one is willing and able to set aside some data that is ``novel," while considering the remaining data to come from the known distribution \cite{digit6}. When combined with a hierarchical classification approach, we can apply novelty detection mechanisms to two different kinds of problems: detecting some entirely new classes (e.g., animals, when the known distribution contains only vehicles), and detecting some classes that are novel at a specific level but not at a broader one, e.g., taxi cabs when the known distribution contains only pickup trucks and sports cars. We will use the term ``novel class'' to refer to both forms of novelty described above, while we will use ``out-of-distribution'' to refer strictly to instances that do not belong to any class in a known taxonomy, no matter how coarse-grained. 

We can formalize these tasks as estimating the probability of membership in one set of classes, with the option of conditioning on membership in another set of classes. 

The standard view of out-of-distribution detection is to estimate the probability that data of fine-grained class $\mathcal{Y}^*$ belongs to the distribution of known fine-grained classes,
$$P(\mathcal{Y}^* \in \mathcal{Y}^{(K)}  | \mathcal{X}).$$

Here, we introduce the problem as \textit{novel class detection}, where we estimate the probability that the fine-grained label of data is in the known children of the parent, conditioned on it being a member of the parent class,
$$P(\mathcal{Y}^* \in c^{(k)}_\text{children}  | \mathcal{Y}^* \in c^{(k)}, \mathcal{X})$$

\noindent where $c^{(k)}_\text{children} = \{c^{(k+1)} | c^{(k+1)} \in c^{(k)}\}$ and child-parent relationships like $\{\text{primate}\} \in \{\text{animal}\}$ hold.

Our description is thus a generalization of the standard view, which reduces to the standard problem when $k = 0$ and we consider $c^{(0)}$ to be the class of all known entities.

\begin{figure*}[ht]
  \centering
    \includegraphics[width=0.8\textwidth]{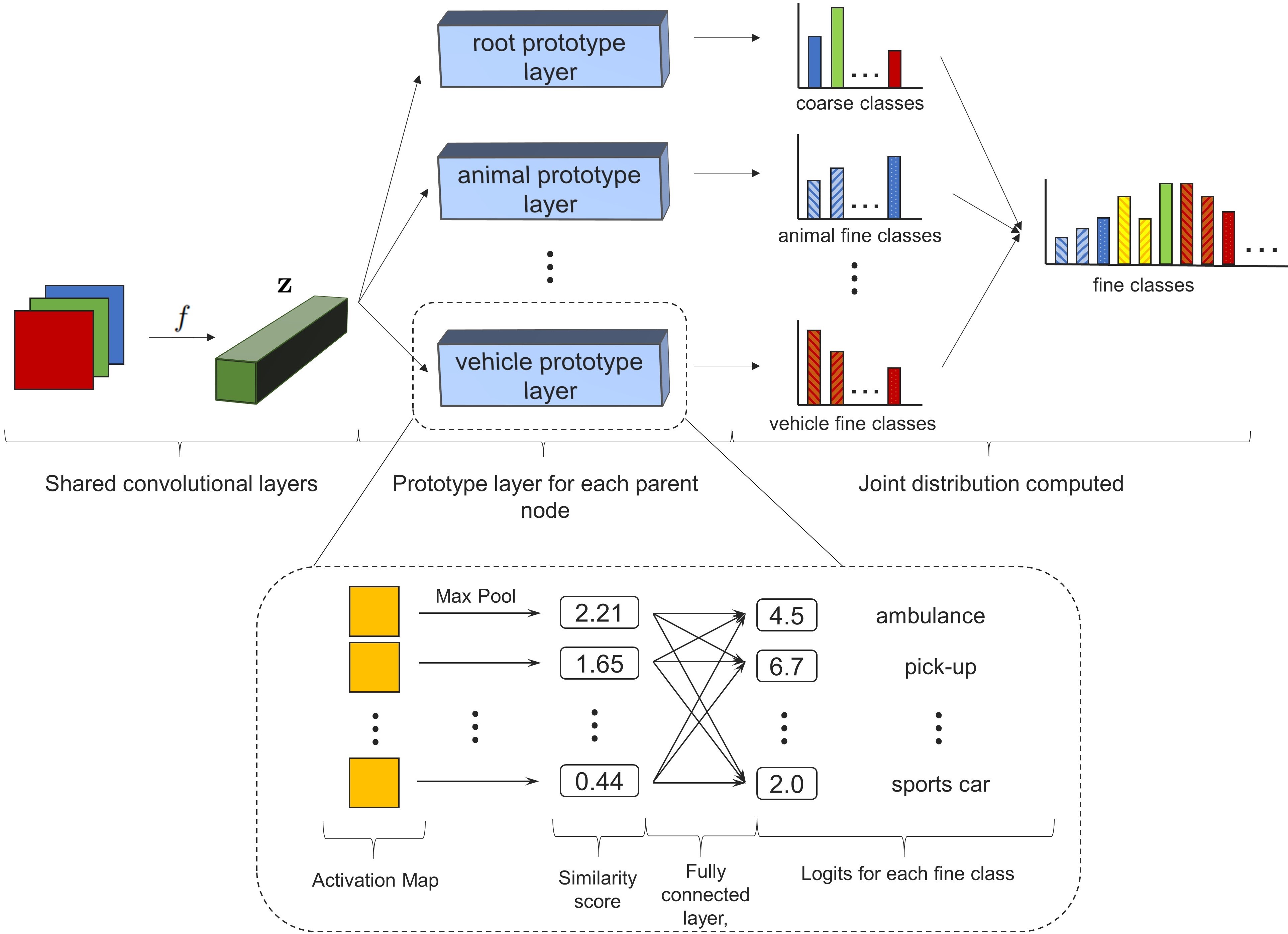}
  \caption{The HPnet architecture.}
  \label{fig:architecture}
\end{figure*}

\section{Model}

In this section we describe our image recognition model and the novel class detection method.

Denote the data as $D = [\mathbf{X}, \mathbf{Y}]= \{(\mathbf{x}_i, \mathbf{y}_i)\}_{i=1}^n$, with hierarchical labels $\mathbf{y}_i \in \mathcal{Y}$ for $i \in \{1, ..., n\}$.

\subsection{Image Recognition Model}

The architecture of our image recognition model is represented in Figure \ref{fig:architecture}. We term our model HPnet, for Hierarchical Prototype network. Our recognition model is a generalization of the model of \citet{thislookslikethat}; with a pre-defined taxonomy consisting of only one level of fine-grained classes, our HPnet model reduces to their model, which we label Pnet. The core components of the Pnet model, including the design of the prototype layer and loss terms, are delineated by \citet{thislookslikethat}, and are described again here for convenience. 

First, a CNN $f$ maps images to a latent space. In our experiments, we use the VGG-16 network \cite{vgg}, with the fully connected layers and classifier removed. We append two 1x1 convolution layers to the end of the network to reduce the dimensionality of the convolutional output from $H \times W \times D$ to $H \times W \times D'$, where $D'\hspace*{-2pt} =\hspace*{-2pt} 32\hspace*{-2pt} < \hspace*{-2pt}D = 512$ and the second activation function is a sigmoid. Let the convolutional output be $\mathbf{z}$. Here, the convolutional output is considered as a set of $HW$ patch vectors each of size $D'$,
$\{\tilde{\mathbf{z}_i}\}_{i=1}^{HW}$. By virtue of the sigmoid activation, the patch vectors are in the unit hypercube in $\mathbb{R}^{D'}$. 

For each parent node in the class taxonomy, there is a prototype layer that operates directly on $\mathbf{z}$. At a high level, prototypes in the latent space will be used to compute feature vectors from the latent representation $\mathbf{z}$. Each element of a feature vector will be a \textit{similarity score} for a particular prototype that will be higher when some patch vector $\tilde{\mathbf{z}} \in \mathbf{z}$ is close to that prototype. 

During training, a set of $m$ prototypes is learned for each prototype layer, denoted as $\mathbf{P}^{c^{(k)}} = \left\{\mathbf{p}_j^{c^{(k)}}\right\}_{j=1}^{m_{c^{(k)}}}$, where $c^{(k)}$ gives the parent class. The prototype layer of class $c^{(k)}$ first transforms $\mathbf{z}$ into a set of $m_{c^{(k)}}$ matrices of size $H \times W$, where the $i^{th}$ matrix is the activation map corresponding to prototype $\mathbf{p}_i^{c^{(k)}}$. Then the max of each activation map is taken to produce a feature vector $\mathbf{v}^{c^{(k)}}$ in $\mathbb{R}^{m_{c^{(k)}}}$. Together, similarity score $j$ of the layer's final feature vector is computed as 
$$
g_{\mathbf{p}_j^{c^{(k)}}}(\mathbf{z}) = \max_{\tilde{\mathbf{z}} \in \text{patches}(\mathbf{z})}
\log\left(1 + 1/(\|\tilde{\mathbf{z}}-\mathbf{p}_j^{c^{(k)}}\|_2^2 + \epsilon)\right).
$$
Finally, a fully connected layer $h^{c^{(k)}}$ transforms the feature vector into a distribution over the classes under that parent node.

There are a few motivations for the design of the prototype layer. By enforcing a constraint that each prototype $\mathbf{p}_j^{c^{(k)}}$ be equal to some patch $\tilde{\mathbf{z}}$ obtained from an image in the training set, we obtain a correspondence between that latent prototype and a visually inspectable receptive field from the training set. Then by upsampling the activation map associated with that prototype to the size of the input image, we get a heat map that shows the localized portions of the input image that highly activate the prototype \cite{thislookslikethat}. Consequently, we can understand a particular classification by checking the prototypes that were highly activated and viewing their corresponding heatmaps overlaid on the input image. 

Further, our prototypes encode for \textit{conditional} information. To give an example, our prototypes for ambulances are used only to distinguish ambulances from other vehicles (here, pickups and sports cars). It is the vehicle prototypes that are used to classify a particular ambulance as a vehicle rather than an animal.

Note that within each set of prototypes $\mathbf{P}^{c^{(k)}}$, we allocate a pre-determined number of prototypes evenly to each child class, so that every child class will be represented in the set of prototypes. The mechanism for this allocation operates in such a way that two prototype vectors can be equivalent, in which case the child node can be considered to have fewer \textit{unique} prototypes than the pre-determined number. In the experiments to follow, we allocate 8 prototypes per child class. We denote the subset of prototypes under this parent layer that are allocated to child $c^{(k+1)}$ as $\mathbf{P}_{c^{(k+1)}}$. 

Lastly, if needed, a distribution over the most fine-grained classes in the taxonomy is obtained by computing the probability of an instance belonging to each class in a path down the taxonomy, for every path down the taxonomy. That is, one computes
\begin{align*}
P(\mathbf{y}_i^{(K)} \hspace*{-3pt}= c^{(K)} &| \mathbf{x}_i) \hspace*{-2pt}
\prod_{k=1}^{K-1} \hspace*{-2pt} P(\mathbf{y}_i^{(k+1)} \hspace*{-3pt}= c^{(k+1)}| \mathbf{y}_i^{(k)} = c^{(k)},\mathbf{x}_i), 
\end{align*}
for each path in the taxonomy, $\{c^{(1)},...,c^{(K)}\}$. 

\subsection{Training Algorithm}

Similar to \citet{thislookslikethat}, we train the model by alternating between optimization of the layers and a projection phase, wherein prototypes are projected onto the closest patches $\tilde{\mathbf{z}}$ in the latent space. This projection phase is necessary since we could not optimize the objective via gradient descent methods while enforcing the constraint that each prototype is always equal to some latent patch from the data. 

Note that \textit{we do not use a VGG-16 model pre-trained on ImageNet}, as this would confound our later novel class detection testing, since our novel class test set also comes from ImageNet. Rather, our VGG-16 is trained with random initialization, and the Pnet and HPnet models are trained with their convolutional layers initialized from the most accurate of the trained VGG-16 models.

\subsubsection{Objective Function.}

The objective function that we aim to minimize is the sum across prototype layers of four terms, a cross entropy between predictions and labels, a clustering term, a separation term, and a regularization term. We use the same clustering and separation terms as \citet{thislookslikethat}, though we adapt them to be specific to each prototype layer. With the set of all parent nodes as $C$, we  minimize
\begin{align*}
&\sum_{c^{(k)} \in C} \bigg[ \sum_{i : \mathbf{y}_i^{(k)} = c^{(k)}} \textrm{CrossEntropy}(h^{c^{(k)}} \circ g_{\mathbf{P}^{c^{(k)}}} \circ f(\mathbf{x}_i), \mathbf{y}_i) \\
&\hspace*{1pt}+\hspace{-1mm} \lambda_1\textrm{Clust}(\mathbf{P}^{c^{(k)}}\hspace*{-5pt}, \mathbf{X}, \hspace*{-2pt}\mathbf{Y}) \hspace*{-3pt}+\hspace*{-3pt} \lambda_2\textrm{Sep}(\mathbf{P}^{c^{(k)}}\hspace*{-5pt}, \mathbf{X}, \hspace*{-2pt}\mathbf{Y})\hspace*{-3pt} +\hspace*{-3pt} \lambda_3 \textrm{Reg}(h^{c^{(k)}})\bigg].
\end{align*}

Each term is explained in turn. The cross entropy encourages accuracy of predictions, and, notably, the sum of the cross entropies over parent classes is equivalent to a single cross entropy between the fine-grained labels of the data and the model's joint distribution over fine-grained classes, since the conditional probabilities decouple through the logarithm.

Let us explain the clustering and separation costs. The clustering cost is designed to encourage the model to map at least one patch vector of each image close to a prototype corresponding to its class. For a given layer, the term is the sum over images of the minimum distance between some patch vector and some prototype of that input image's class.
\begin{align*}
\textrm{Clust}(&\mathbf{P}^{c^{(k)}}, \mathbf{X}, \mathbf{Y})
= \\ &\sum_{i : \mathbf{y}_i^{(k)} = c^{(k)}} \min_{j: \mathbf{p}_j \in \mathbf{P}_{c_i^{(k+1)}}} \min_{\tilde{\mathbf{z}} \in \text{patches}(f(\mathbf{x}_i))} \|\tilde{\mathbf{z}} - \mathbf{p}_j\|_2^2.
\end{align*}

The separation cost is designed to encourage the model to avoid mapping any patch vectors of an image close to a prototype corresponding to a \textit{different} class. For a given layer, the term is the negative sum over images of the minimum distance between some patch vector and some prototype \textit{not} belonging to that input image's class.
\begin{align*}
\textrm{Sep}(&\mathbf{P}^{c^{(k)}}, \mathbf{X}, \mathbf{Y})
= \\ &-\sum_{i : \mathbf{y}_i^{(k)} = c^{(k)}} \min_{j: \mathbf{p}_j \not\in \mathbf{P}_{c_i^{(k+1)}}} \min_{\tilde{\mathbf{z}} \in \text{patches}(f(\mathbf{x}_i))} \|\tilde{\mathbf{z}} - \mathbf{p}_j\|_2^2.
\end{align*}
These additional cost terms induce a clustering structure in the latent space, ensuring that prototypes encode for information that is specific to the class they correspond to. We confirm this empirically by checking which patch vectors are close to each prototype, as elaborated on in Section \ref{section:interpretation}.

Lastly, Reg is a regularization term on the fully connected layers of each prototype layer. With respect to a given class, the term imposes $l_2$ regularization on weights that connect to similarity scores of prototypes belonging to that class, while imposing $l_1$ regularization on the weights that connect to similarity scores of prototypes belonging to \textit{other} classes. Thus as the model tallies the evidence for an instance belonging to a certain class, the Reg term encourages the model to rely \textit{only} on similarity scores of prototypes belonging to that class, as the weights connecting to scores of other class's prototypes will be sparse. This greatly simplifies model interpretation. 

We give an approximate algorithm for minimizing the objective function. The algorithm proceeds through several training phases that are described below. The number of epochs spent in each phase is given in Table \ref{table:epochs} in Appendix \ref{appendix}.

\begin{figure*}[ht]
  \centering
    \includegraphics[width=0.8\textwidth]{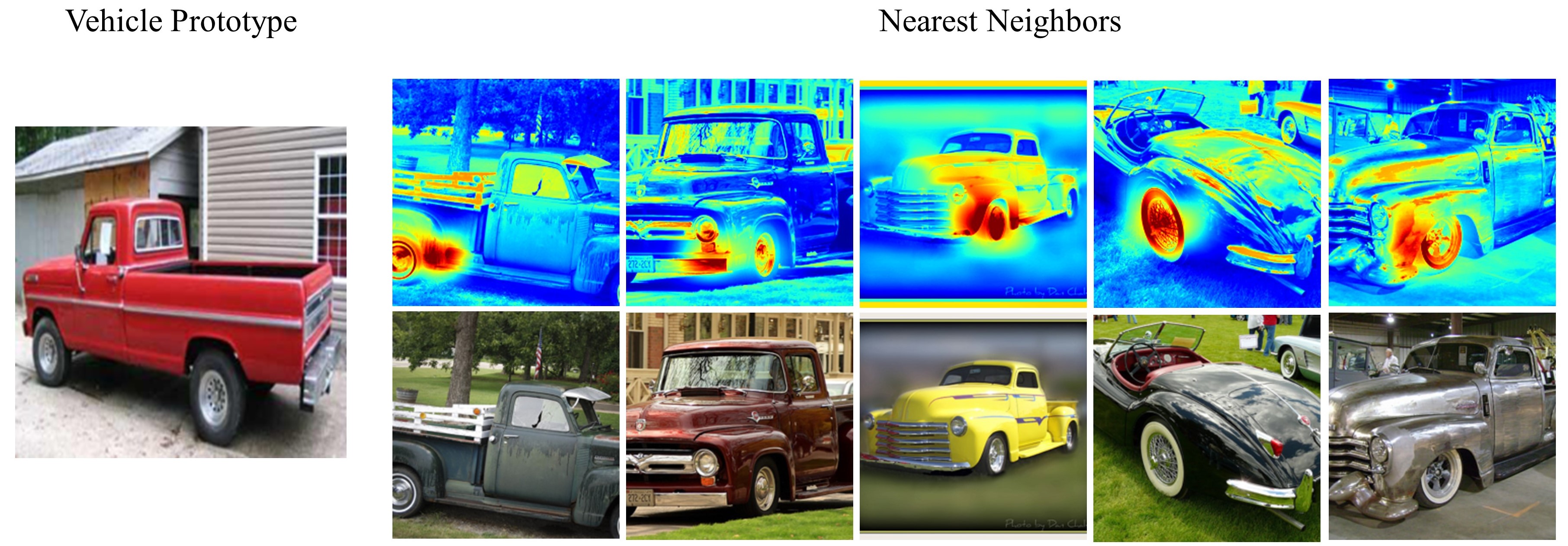}
  \caption{The 5 nearest neighbors for this vehicle prototype from the test set. From the neighbors, it appears that this prototype encodes for wheels and wheel wells (the wheel edges). Notice that this prototype encodes for these properties in a \textit{vehicle-general} manner. Among the 5 nearest neighbors, the wheels of both pickups and sports cars are activated.}
  \label{fig:knn}
\end{figure*}

\subsubsection{Convolutional Layers and Prototypes.}

In the first optimization phase, the objective is optimized with respect to the weights of $f$ and each $\mathbf{P}^{c^{(k)}}$ via stochastic gradient descent, while each $h^{c^{(k)}}$ remains fixed.

To initialize the weights of each $h^{c^{(k)}}$ layer, we adopt the technique of \citet{thislookslikethat}, which is to set class connections to 1 when they correspond to similarity scores of prototypes belonging to that class, and $-$.5 otherwise. That is, supposing that the logit for a particular class $c$ with parent $c^{(k)}$ is computed as $\alpha_c^T\mathbf{v}^{c^{(k)}}$, we set the $j^{th}$ element of $\alpha_c$ to 1 if $\mathbf{p}_j^{c^{(k)}}$ is a prototype allocated to class $c$ and to $-$.5 if the prototype was allocated to another class. Thus evidence for an image belonging to a certain class accrues as it activates prototypes belonging to that class and diminishes as it activates prototypes belonging to other classes. 

\subsubsection{Optimization of All Layers.}

In this phase, we optimize all layers of the network at once, including the fully connected layers in each prototype layer. It is in this training phase that the weights of each $h^{c^{(k)}}$ layer become sparse. 

\subsubsection{Projection of Prototypes.} \label{projection}

Every five epochs of the above two phases, we project prototypes onto the patch vectors from the training data that they are closest to, with the constraint that prototypes can only be projected onto patch vectors from instances belonging to the class to which the prototypes have been pre-allocated. Since we do not restrict distinct prototypes from being projected onto the same patch vector, multiple prototype vectors may be equal to each other after the projection. This phase is necessary to achieve a direct correspondence between prototypes and latent representations from the training data.

\subsubsection{Convex Optimization of Fully Connected Layers.}

Following each projection phase, we perform the same convex optimization described by \citet{thislookslikethat}, but now we do so for the fully connected layer in each prototype layer.

\subsection{Novel Class Detection Model} \label{novel class detection}

Since we are treating a problem not considered by \citet{digit6}, we must adapt the model-fitting scheme to the problem at hand. 
We fit a novel class detector for every parent node to discriminate between children of that node from the known distribution and not-yet-seen children of that node. For instance, a model is fit on a dataset consisting of vehicles seen during training of the image recognition model, vehicles = \{ambulance, sports car, pickup\}, as well as novel vehicles not seen during training of the recognition model, $\text{vehicles}^*$ = \{cab, forklift, tractor, mountain bike\}. The goal is to discriminate between these sets. We test a number of classifiers from \citet{digit6}. 

By fitting a probabilistic classifier for each parent node, we are able to produce nuanced predictions such as ``novel vehicle", which follows formally from the highest predicted probability being 
\begin{align*}
P(&\mathbf{y^*} \notin \text{vehicles}, c^{(1)} = \text{vehicle}| \mathbf{x}) =\\
&P(\mathbf{y^*} \notin \text{vehicles} | c^{(1)} = \text{vehicle}, \mathbf{x}) P(c^{(1)} = \text{vehicle} | \mathbf{x})
\end{align*}
where the left-hand probability is obtained from the novel vehicle detector and the right-hand probability is obtained from the image recognition model.

\section{Experiments}

For a model trained on a subset of ImageNet data, we interpret a model prototype, show a case study of the model classifying a novel image, and give image recognition and novel class detection accuracies.

For training our recognition model, we select a subset of 15 ImageNet classes and define a taxonomy over them as shown in Figure \ref{fig:taxonomy}. Note that we hold out 50 images from the 1300 training images for each class to create a validation set used for early stopping in training. The test accuracies we report are calculated only after training is complete.

For later novel class detection, we hold out 15 additional classes that fall under the same coarse classes in $\mathcal{Y}^{(1)}$, though in the novel data we have four kinds of vehicles and no scuba divers. The taxonomy for this data is shown in Figure \ref{fig:novel_taxonomy} in Appendix \ref{appendix}.

\subsubsection{Data Augmentation.}

We implement the CEDA data augmentation technique of \citet{calibration}. This technique involves including with every training batch an equal number of random noise images. The only loss term these images are implicated in is a cross entropy between their predicted class distributions and a uniform distribution over classes, thus encouraging the model to be maximally uncertain over random noise. The CEDA technique may improve the clustering quality of the HPnet latent space (see Table \ref{table:knn_stats}), and it does not lower model accuracy \cite{calibration}.

Besides this technique, we use the standard ImageNet cropping procedure of, for training, random resized crops of size 224 by 224 and, for testing, resizing to 256 by 256 then cropping to 224 by 224.

\begin{table}
	\centering
	\begin{tabular}{lSS}
		\toprule
		& \multicolumn{2}{c}{\% Correct Neighbors} \\ \cmidrule(lr){2-3}
		Model & {Train} & {Test} \\
		\midrule
		HPnet & 79.24 & 76.2 \\
		HPnet + CEDA & 84.90 & 79.24 \\
		\bottomrule
	\end{tabular}
	\caption{(Latent-space clustering quality.) Proportion of the nearest neighbors to the prototypes that belong to the same class as their neighboring prototype. Prototypes tend to be surrounded by patch vectors from images that belong to the correct classes, indicating a learned clustering quality to the latent space. Note that these two models are initialized with VGG-16 base models that achieve differing accuracy, so the change in clustering quality could result from the differing initializations rather than the CEDA technique. We give confidence intervals in Table \ref{table:knn_CIs} in Appendix \ref{appendix}.}
	\label{table:knn_stats}
\end{table}

\subsection{Interpreting the Recognition Model's Latent Space} \label{section:interpretation}

How do we interpret the features learned by the model? We can inspect each prototype along with the images of patch vectors that are closest to it in the latent space. In Figure \ref{fig:knn}, we show a vehicle prototype, the test images whose patch vectors are closest to the prototype in the latent space, and those images with prototype activation maps overlaid on them. Notice that this prototype encodes for wheels in a \textit{vehicle-general} manner. Among the 5 nearest neighbors, the wheels of both pickups and sports cars are activated.

We would also like to capture a global perspective on the clustering quality of the latent space in order to check that prototypes tend to be surrounded by patch vectors from images of the same class. To do so, for each prototype we compute the percentage of its five nearest neighbors that belong to the correct class (e.g., a vehicle for one of the vehicle prototypes); we average these percentages to obtain a metric of clustering quality, which we show in Table \ref{table:knn_stats}.

\subsection{Classifying a Test Image}

\begin{figure}[t]
  \centering
    \includegraphics[width=0.4\textwidth]{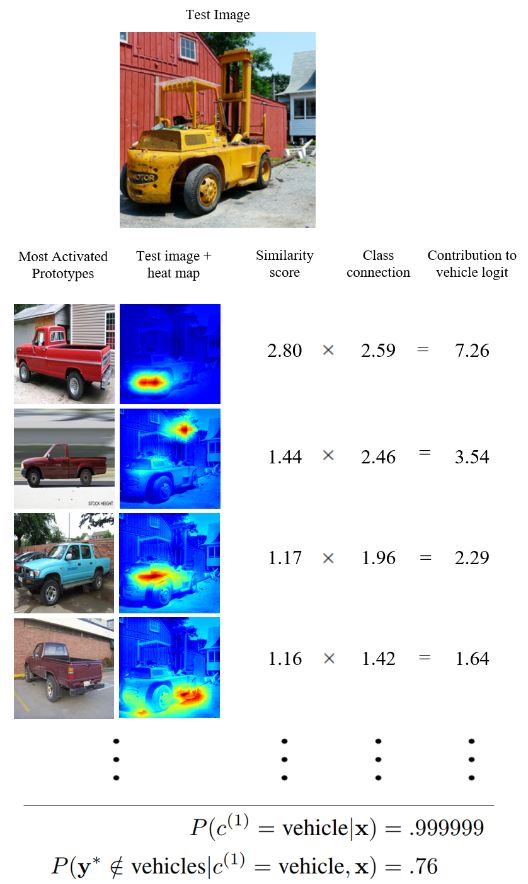}
  \caption{A forklift is classified as a novel vehicle. Note that the strongest evidence for this forklift being a vehicle is its possession of a wheel, as evidenced by the most activated prototype. For context, the image recognition model used here obtained 61\% accuracy on classifying the novel forklifts as vehicles, and the logistic model for novel class detection obtained 69\% test accuracy on discriminating familiar vehicles from novel vehicles. These top four prototypes accounted for 74\% of the magnitude of the vehicle logit.}
  \label{fig:case_study1}
\end{figure}

In this section, we give an example of a forklift being classified as a novel vehicle. A diagram is shown in Figure \ref{fig:case_study1}. By relying on learned concepts such as wheels, the model successfully classifies the forklift as a vehicle. The novel vehicle detector also successfully classifies it as a novel vehicle, i.e. \textit{not} in the set of \{ambulance, pickup, sports car\}. For additional examples, see Figures \ref{fig:case_study3} and \ref{fig:case_study2} in Appendix \ref{appendix}. 

\subsection{Image Classification Accuracy}
\begin{table}[h]
  \centering
  \begin{tabular}{*{4}{l}}
    \toprule
    & \multicolumn{3}{c}{Test Accuracies by Model} \\
    \cmidrule(lr){2-4}
    Model & F-ID & C-ID & C-Novel \\
    \midrule
	VGG-16 + CEDA & 82.19 & 92.83 & 62.31 \\
	Pnet + CEDA & 81.60 & 92.56 & 60.17 \\
	HPnet + CEDA & 82.61 & 93.57 & 62.16 \\
    \bottomrule
  \end{tabular}
  \caption{(Test accuracies for each model.) The accuracies include the fine-grained accuracy on in-distribution data (F-ID), the coarse-grained accuracy on in-distribution data (C-ID), and the coarse-grained accuracy on novel data (C-Novel). Accuracies are averages across five models per method. We give confidence intervals in Table \ref{table:accuracy_CIs} in Appendix \ref{appendix}. Note that the weights of the convolutional layers for the Pnet and HPnet models are always initialized from the trained weights of the most accurate VGG-16 model, which achieves 84.93\% F-ID accuracy.}
  \label{table:accuracy_models}
\end{table}

We give the test accuracies for each model in Table \ref{table:accuracy_models}. The models include the VGG-16 network, our HPnet, and a flat version of our model, Pnet. See Appendix \ref{appendix} for the method of computing coarse-grained predicted probabilities for the VGG-16 network. 

Our model attains on average similar accuracy to its black-box and flat counterparts. However, relative to the pre-trained VGG-16 model used to initialize HPnet, there is an average drop in fine-grained accuracy of 2.32\%. This drop is to be expected given the additional constraints introduced for purposes of interpretability; it remains to practitioners to assess if a gain in accuracy of this size is worth sacrificing interpretability for. This gap will also likely narrow as the space of training approaches for prototype-based models like HPnet is explored to the extent that it has been for the VGG class of models.

Finally, we observe that it is possible to correctly classify novel data at the coarse level at far above the chance rate. Random performance on the novel data would yield about $17\%$ accuracy.

\subsection{Novel Class Detection}
We test three methods from \citet{digit6}, ScoreSVM, PbThreshold, and a logistic regression. The first of these is an SVM (with linear kernel) on the model logits, the second is a simple threshold on the max predicted probability, and the last is a logistic regression on the model logits. We apply each of these three methods to two image recognition models, one with CEDA and one without, for a total of six methods.

The accuracies obtained are in line with those reported by \citet{digit6}, and there is no clearly superior method. The logistic regression for the HPnet + CEDA model, which is used in Figure \ref{fig:case_study1}, obtains $73.72\%$ accuracy on average over five recognition models. We give the accuracies with confidence intervals in Table \ref{table:novel} in Appendix \ref{appendix}, where the testing procedure is described as well.   

\section{Conclusion}

We provide a model that classifies objects at each level in a semantic taxonomy, identifies when objects are novel at fine-grained levels of its taxonomy, and uses directly interpretable features that are tailored to each parent node in the taxonomy. In general, this is the first vision model to accomplish all three tasks in a synchronized manner. In particular, it is the first CNN-based model to organize prototypes hierarchically and the first to perform novel class detection across levels of the class taxonomy. As a result, the explanations are much richer, leading to dramatically improved interpretability. An application to ImageNet demonstrates the viability of the method. 

\section{Acknowledgements}

The authors would like to thank Alina Barnett and Chaofan Tao at Duke University, as well as Jonathan Su at MIT Lincoln Laboratory, for their helpful contributions and feedback on this work.

\appendix

\section{Appendix} \label{appendix}

\subsection{Coarse Predictions from Flat Models}
There are two ways to compute coarse-grained predictions from a flat model, as is needed to obtain the C-Novel metric for the VGG and Pnet models in Table \ref{table:accuracy_models}. First, one can take the fine-grained prediction of the model and check if it is in the same coarse class as the true label (e.g., predicting the fine class to be ambulance yields the correct coarse prediction if the true label is sports car). Alternatively, one could sum up the predicted probabilities over the fine-grained classes within each coarse class to get predicted coarse class probabilities, and then check that the max probability is for the true coarse class (e.g., $p(\text{vehicle})$ is obtained as $p(\text{ambulance}) + p(\text{pickup}) + p(\text{sports car})$, and the coarse prediction for an instance of vehicle is only correct if this $p(\text{vehicle})$ is the max of these coarse-grained probabilities). Empirically, we observe that the second approach involving sums over fine-grained probabilities outperforms the first approach, so we present accuracies computed using the second approach. 

\subsection{Testing Procedure for Novel Class Detection}
We present accuracies of novel class detection methods in Table \ref{table:novel}. Accuracies are averaged across models fit to logits from five image recognition models. An accuracy metric for a single recognition model is obtained as follows: for each parent node, a test accuracy is obtained by an average across leave-one-class-out tests. Thus for the vehicle detector, four tests are performed, where each class in \{cab, forklift, tractor, mountain bike\} serves as the test novel dataset once, and each time the remaining three novel classes are included in the training data. This split is justified in \cite{digit6}, since the classes in the test data should not have been seen by the novel class detector at all during training. Training data are drawn from the training partition of the ImageNet data, while testing data are drawn from the testing partition. Data are always balanced to consist of half familiar classes and half novel classes. Hyperparameters for the models are tuned with 50 held out images from the training data.

\subsection{Tables and Figures}

\begin{figure}[h]
  \centering
  \includegraphics[width=0.45\textwidth]{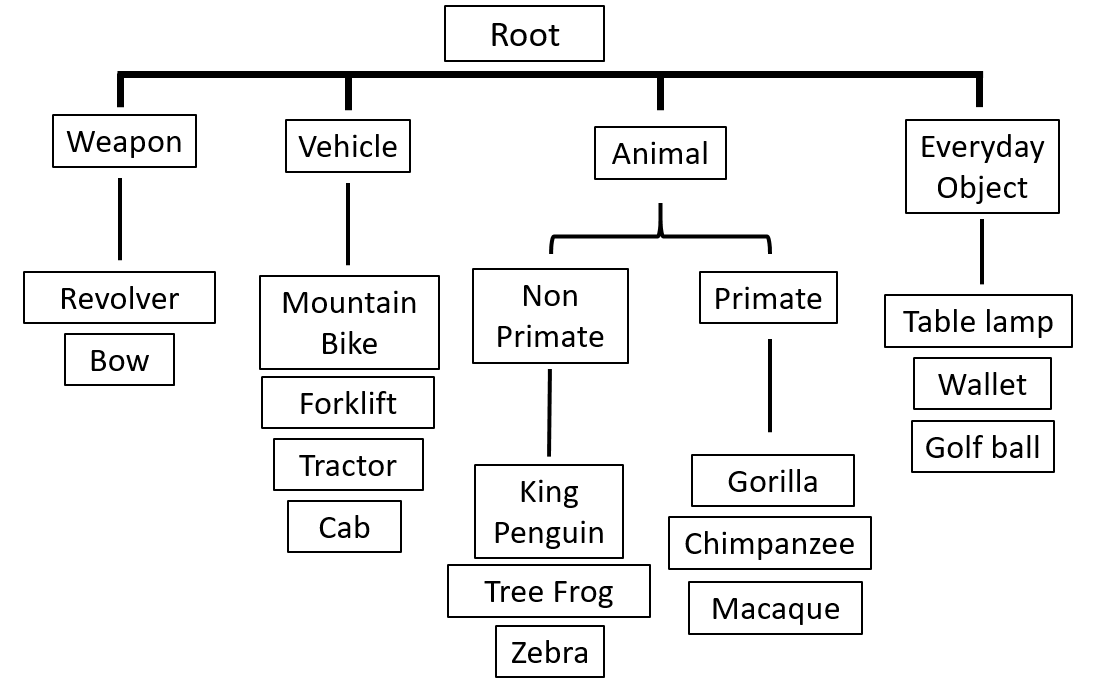}
  \caption{The novel data taxonomy. The image recognition model does not see these data during training.}
  \label{fig:novel_taxonomy}
\end{figure}

\begin{table}[h]
	\centering
	\begin{tabular}{ll}
		\toprule
		Phase & {Num. Epochs} \\
		\midrule
		Conv. Layers & $5$ \\
		All Layers & $45$ \\
		Convex Opt. & $2$ \\
		\bottomrule
	\end{tabular}
	\vspace{3mm}
	\caption{(Number of epochs spent in each training phase.) The projection phase occurs every 5 epochs of the ``Conv. Layers'' and ``All Layers'' phases. Projection phases are always followed by convex optimization phases. The very last convex optimization optimization phase lasts for 10 epochs rather than 2.}
	\label{table:epochs}
\end{table}

\begin{table}[h]
	\centering
	\begin{tabular}{lll}
		\toprule
		& \multicolumn{2}{c}{\% Correct Neighbors} \\ \cmidrule(lr){2-3}
		Model & {Train} & {Test} \\
		\midrule
		HPnet & $79.24 \pm 1.65$ & $76.20 \pm 1.56$ \\
		HPnet + CEDA & $84.90 \pm 2.2$ & $79.24 \pm 1.83$ \\
		\bottomrule
	\end{tabular}
	\caption{(Latent-space clustering quality.) Proportion of the nearest neighbors to the prototypes that belong to the same class as their neighboring prototype. We report the average proportion across five trained models for each method, as well as the 95\% confidence intervals.}
	\label{table:knn_CIs}
\end{table}

\begin{figure*}[h]
  \centering
    \includegraphics[width=0.7\textwidth]{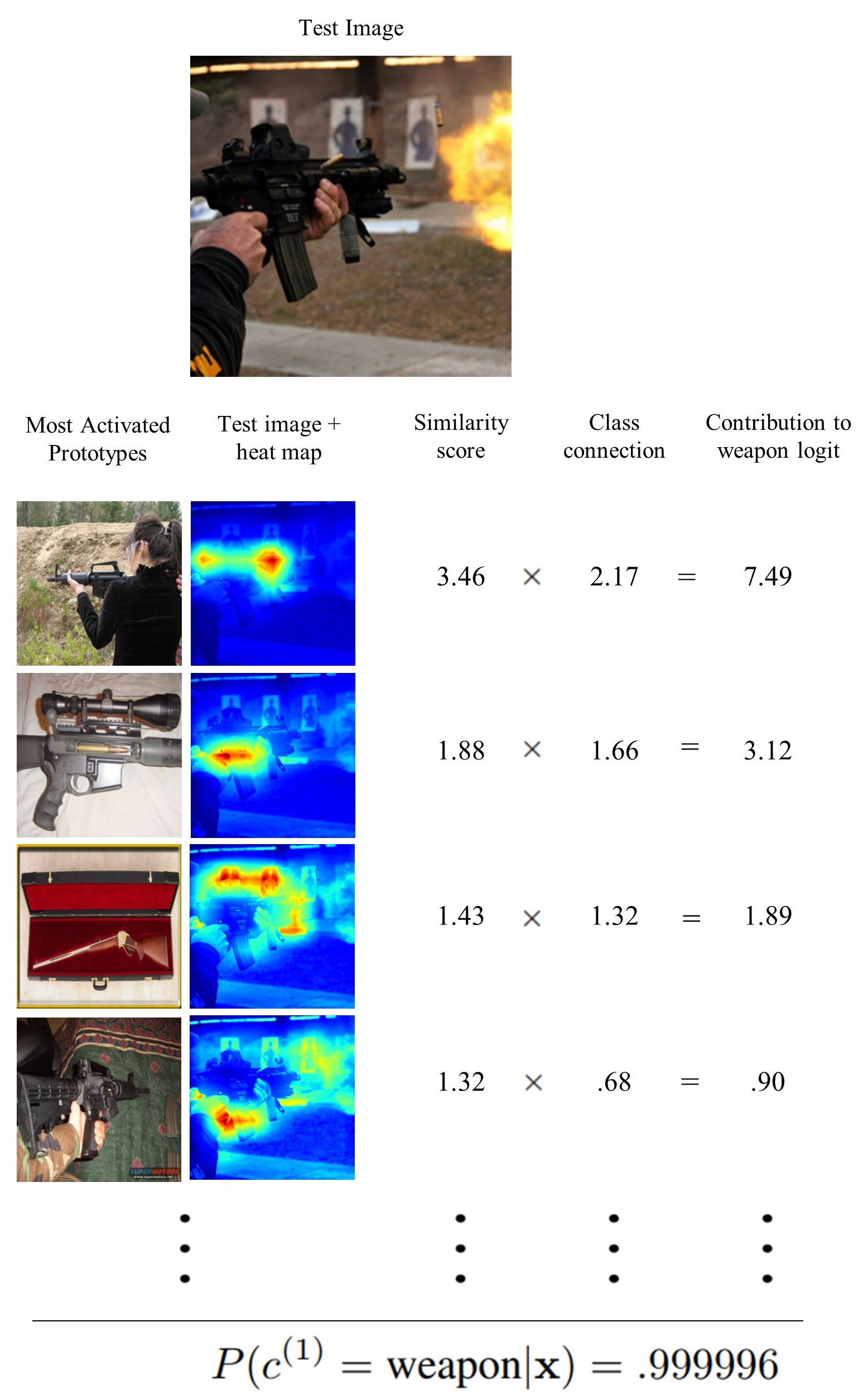}
  \caption{An assault rifle is classified as a weapon. The second most activated prototype detected the weapon's trigger (see Figure \ref{fig:weapon_knn} in Appendix \ref{appendix}), and the fourth most activated prototype detected the hand holding the weapon. Together, the top four prototypes account for 75\% of the magnitude of the weapon logit. For context, the model obtained 83\% accuracy on the weapon class.}
  \label{fig:case_study3}
\end{figure*}

\begin{figure*}[h]
  \centering
    \includegraphics[width=0.7\textwidth]{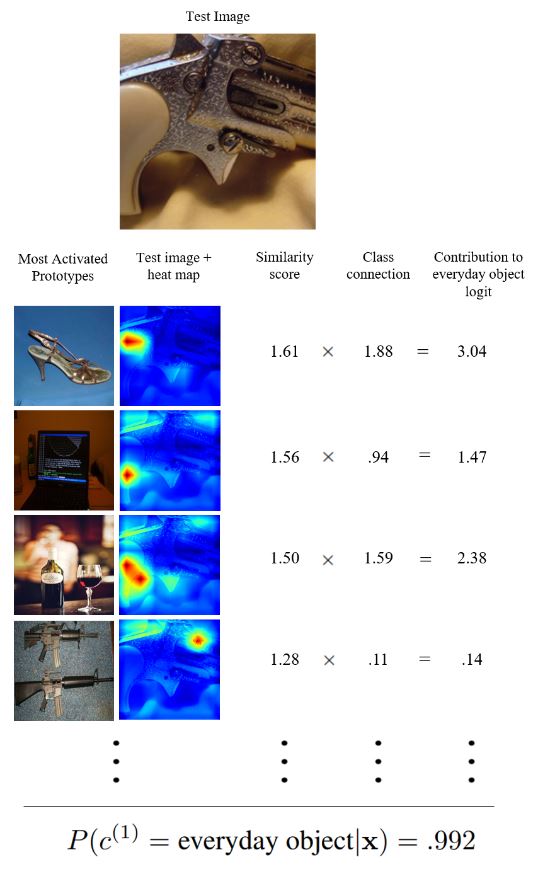}
  \caption{A revolver is classified as an everyday object. For context, the image recognition model obtained 53\% accuracy on classifying the novel revolvers as weapons. The availability of the prototypes here is helpful for diagnosing what went wrong in the prediction; an investigation of each prototypes' nearest neighbors would help reveal what lead to the mistake. With this information, one gains an idea as to what data collection or augmentation would help prevent this kind of model error.}
  \label{fig:case_study2}
\end{figure*}

\begin{figure*}[h]
  \centering
    \includegraphics[width=0.8\textwidth]{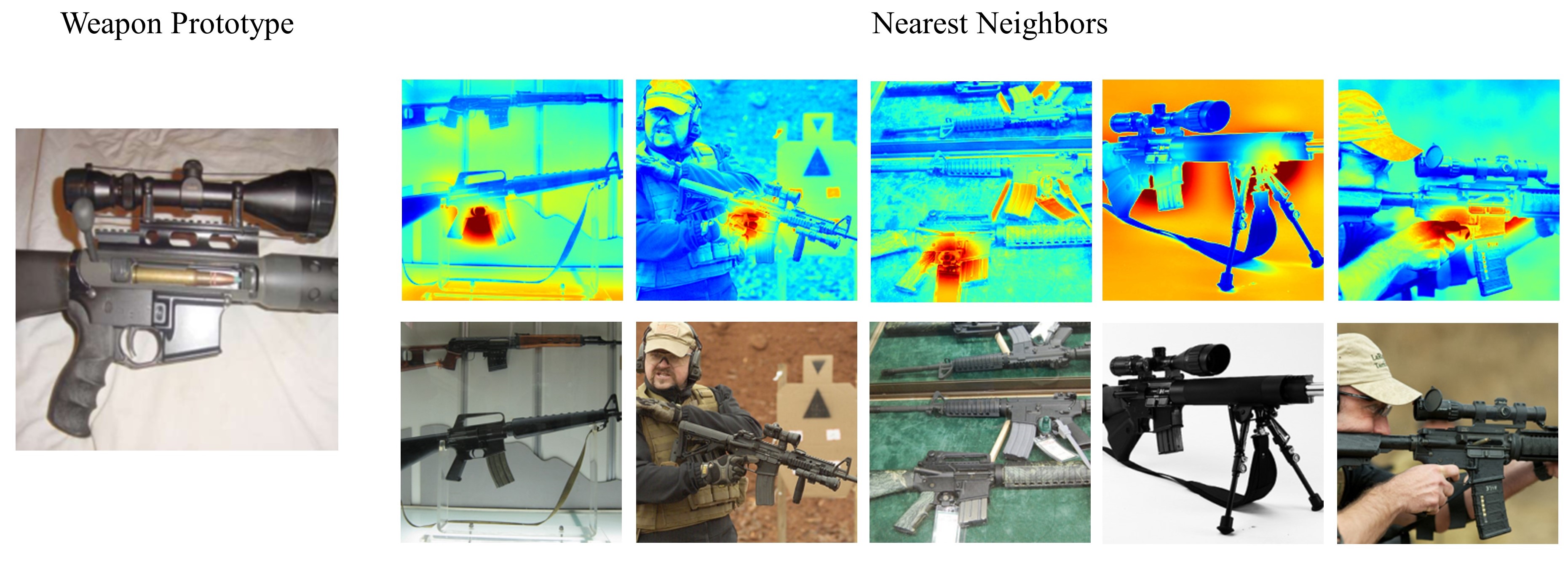}
  \caption{One model's 5 nearest neighbors for this weapon prototype from the test dataset. The learned prototype encodes for the presence of a trigger and the trigger guard, or a similar pattern appearing in the fourth nearest neighbor.} 
  \label{fig:weapon_knn}
\end{figure*}

\begin{figure*}[h]
  \centering
    \includegraphics[width=0.8\textwidth]{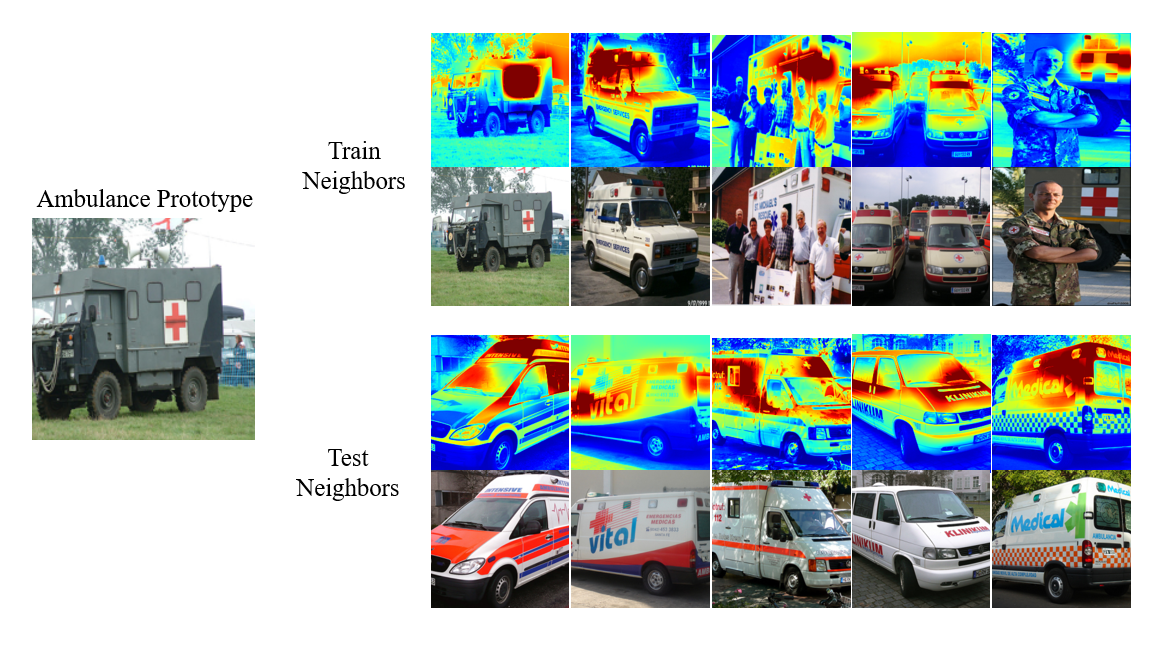}
  \caption{One model's 5 nearest neighbors for this ambulance prototype from the train and test datasets. The learned prototype encodes for a red-on-white pattern exemplified in the training nearest neighbors, or, judging by the last two neighbors from the test data, a more general color-on-white pattern.} 
  \label{fig:ambulance_knn}
\end{figure*}

\begin{figure*}[h]
  \centering
    \includegraphics[width=0.9\textwidth]{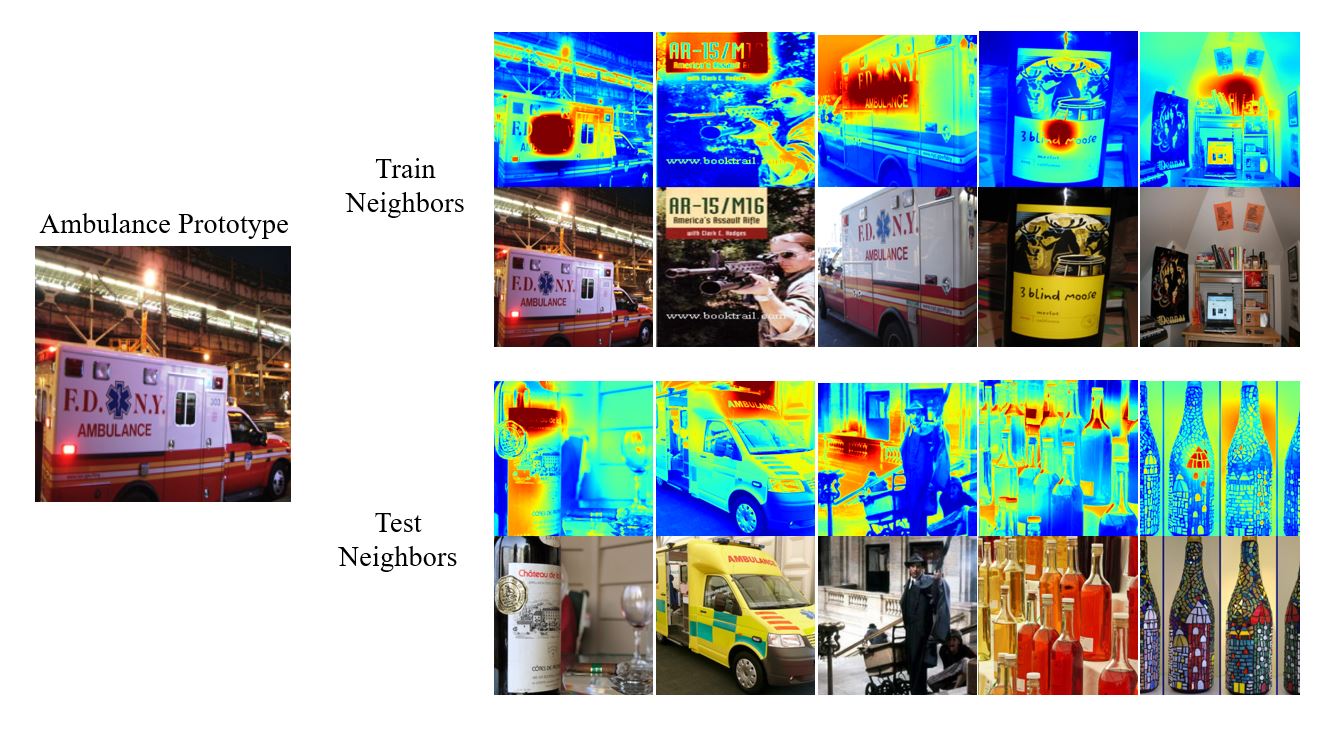}
  \caption{One model's 5 nearest neighbors for this ambulance prototype from the train and test datasets. The learned prototype encodes for \textit{text} of variance kinds, or, judging by the some of the nearest test neighbors, a dark-lines-on-light-background pattern. While the prototype is highly activated by the assault rifle and wine bottle images here, it is possible that ambulances that highly activate this prototype are never classified as assault rifles or wine bottles, by virtue of always being classified as vehicles to begin with, rather than weapons or everyday objects.} 
  \label{fig:ambulance_knn2}
\end{figure*}

\clearpage

\begin{table}[h]
  \centering
  \begin{tabular}{*{4}{l}}
    \toprule
    & \multicolumn{3}{c}{Test Accuracies by Model (w/ CEDA)} \\
    \cmidrule(lr){2-4}
    Model & F-ID & C-ID & C-Novel \\
    \midrule
	VGG-16 & $82.19 \pm 1.53$ & $92.83 \pm .78$ & $62.31 \pm 1.72$ \\
	Pnet & $80.67 \pm .78$ & $91.79 \pm .45$ & $60.58 \pm .32$\\
	HPnet & $82.61 \pm .44$ & $93.57 \pm .43$ & $62.16 \pm .70$ \\
    \bottomrule
  \end{tabular}
  \caption{(Test accuracies for each model.) The accuracies include the fine-grained accuracy on in-distribution data (F-ID), the coarse-grained accuracy on in-distribution data (C-ID), and the coarse-grained accuracy on novel data (C-Novel). The accuracies reported are averages across five trained models per method, with 95\% confidence intervals. Our interpretable model, HPnet, performs similarly to its black-box counterpart VGG-16 as well as its flat counterpart Pnet. Note that the weights of the convolutional layers for the Pnet and HPnet models are always initialized from the trained weights of the best VGG-16 model, which achieves 84.93\% F-ID accuracy.}
  \label{table:accuracy_CIs}
\end{table}

\begin{table}[h]
	\centering
	\begin{tabular}{ll}
		\toprule
		Classifier & {Accuracy} \\
		\midrule
		PbThreshold & $52.05 \pm 1.86$ \\
		PbThreshold + CEDA & $51.22 \pm 1.50$ \\
		ScoreSVM & $74.69 \pm .96$ \\
		ScoreSVM + CEDA & $73.94 \pm 1.13$ \\
		Logistic Reg. & $74.75 \pm 1.05$ \\
		Logistic Reg. + CEDA & $73.72 \pm .95$ \\
		\bottomrule
	\end{tabular}
	\vspace{3mm}
	\caption{(Novel class detection accuracies.) Accuracies are averaged across five models per method. In-distribution and novel data are balanced in testing. For each method, the test detection accuracies are averaged across parent nodes.}
	\label{table:novel}
\end{table}

\clearpage
\bibliography{main}
\bibliographystyle{aaai}

\end{document}